\documentclass[conference]{IEEEtran}
\usepackage[utf8]{inputenc}
\IEEEoverridecommandlockouts
\usepackage{cite}
\usepackage{amsmath,amssymb,amsfonts}
\usepackage{graphicx}
\usepackage{textcomp}
\def\BibTeX{{\rm B\kern-.05em{\sc i\kern-.025em b}\kern-.08em
    T\kern-.1667em\lower.7ex\hbox{E}\kern-.125emX}}
    
\usepackage{algorithm,algpseudocode}

\usepackage{mathtools, nccmath}
\DeclarePairedDelimiter{\abs}\lvert\rvert
\usepackage{hyperref}
\usepackage{tikz}
\newcommand*\circled[1]{\tikz[baseline=(char.base)]{
            \node[shape=circle, draw=black!60, fill=black!, text=white, inner sep=1pt] (char) {#1};}}

\usepackage{multirow}
\usepackage[export]{adjustbox}

\newcount\Comments  
\newcommand{\multiline}[1]{%
  \begin{tabularx}{\dimexpr\linewidth-\ALG@thistlm}[t]{@{}X@{}}
    #1
  \end{tabularx}
}
\newcommand{\prof}[1]{\ifnum\Comments=1\textcolor{red}{#1}\fi}
\usepackage{subcaption}
\begin{document}

\title{CRISP: Hybrid Structured Sparsity for\\ Class-aware Model Pruning
}

\author{
    \IEEEauthorblockN{Shivam Aggarwal\textsuperscript{*}, Kuluhan Binici\textsuperscript{* $\dagger$},
    Tulika Mitra\textsuperscript{*}}
    \IEEEauthorblockA{\IEEEauthorrefmark{1}School of Computing, National University of Singapore}
    \IEEEauthorblockA{\IEEEauthorrefmark{ 2}Institute for Infocomm Research, A*STAR, Singapore\\
    \{shivam, kuluhan, tulika\}@comp.nus.edu.sg}
}


\maketitle

\begin{abstract}
Machine learning pipelines for classification tasks often train a universal model to achieve accuracy across a broad range of classes. However, a typical user encounters only a limited selection of classes regularly. This disparity provides an opportunity to enhance computational efficiency by tailoring models to focus on user-specific classes. Existing works rely on unstructured pruning, which introduces randomly distributed non-zero values in the model, making it unsuitable for hardware acceleration. Alternatively, some approaches employ structured pruning, such as channel pruning, but these tend to provide only minimal compression and may lead to reduced model accuracy. In this work, we propose CRISP, a novel pruning framework leveraging a hybrid structured sparsity pattern that combines both fine-grained N:M structured sparsity and coarse-grained block sparsity. Our pruning strategy is guided by a gradient-based class-aware saliency score, allowing us to retain weights crucial for user-specific classes. CRISP achieves high accuracy with minimal memory consumption for popular models like ResNet-50, VGG-16, and MobileNetV2 on ImageNet and CIFAR-100 datasets. Moreover, CRISP delivers up to 14$\times$ reduction in latency and energy consumption compared to existing pruning methods while maintaining comparable accuracy. Our code is available \href{https://github.com/shivmgg/CRISP/tree/main}{here}. 
\end{abstract}


\section{Introduction}
Machine learning has become ubiquitous for a broad range of applications, from computer vision \cite{vgg16} to natural language processing \cite{nlp}. While universal models trained on large-scale datasets provide accurate predictions over a wide range of classes, individual users often encounter only a tiny subset of these classes in their daily lives \cite{9218741}. As such, deploying these models to end-users can be resource-intensive and wasteful. One way to address this challenge is to personalize these models for individual users, focusing only on the classes relevant to them.

Class-aware personalization offers a promising avenue to significantly reduce the resource requirements of deployed models, thereby enhancing their efficiency and reducing the carbon footprint on edge devices. One naive approach for this task is unstructured pruning, where weights can be randomly pruned based on class preferences. However, such methods do not yield noticeable hardware improvements unless the model is highly sparse ($\sim$99\%) due to the irregular memory access pattern ~\cite{9065523}. Existing class-aware pruning methods such as CAP'NN \cite{9218741}, MyML \cite{DBLP:conf/dac/GoyalBD21} and OCAP \cite{OCAP} leverage structured pruning such as channel or filter pruning based on user affinity, but they struggle to maintain high model accuracy, particularly at high sparsity levels as they eliminate entire channels from the weight matrix.

\begin{figure}[!h]
\begin{minipage}[t]{0.48\linewidth}
    \includegraphics[width=\linewidth]{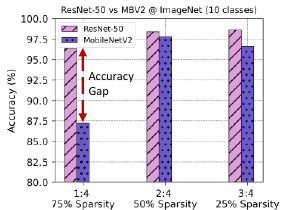}
    \caption{Models at different N:M ratios}
    \label{fig:diff_nm}
\end{minipage}%
    \hfill%
\begin{minipage}[t]{0.48\linewidth}
    \includegraphics[width=\linewidth]{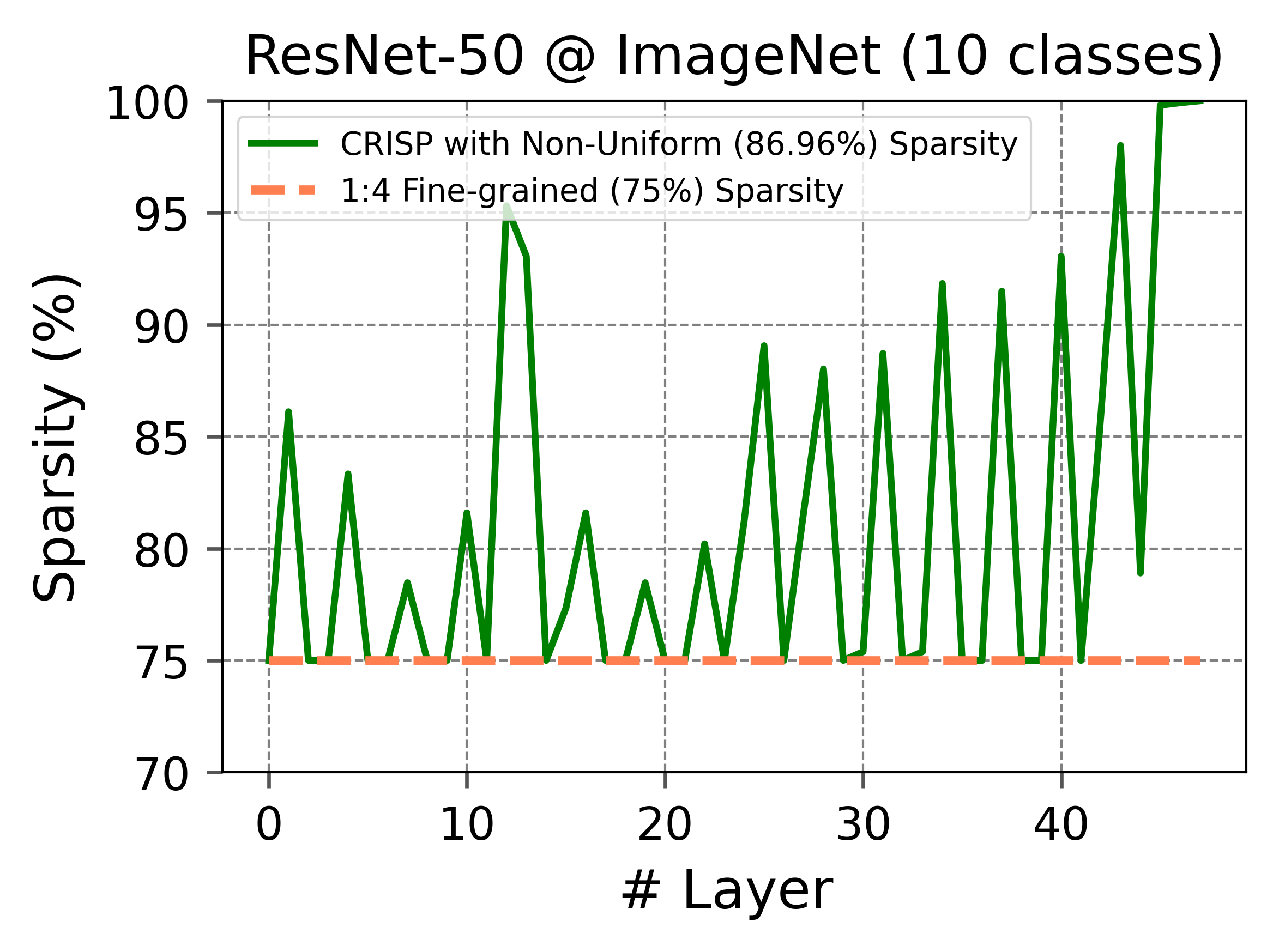}
    \caption{Layer-wise sparsity distribution}
    \label{fig:layerwise_spar}
\end{minipage} 
  \vspace{-5mm}
 \label{fig:group}
\end{figure}

\begin{figure}[!h]
	\centering	\includegraphics[width=0.95\linewidth]{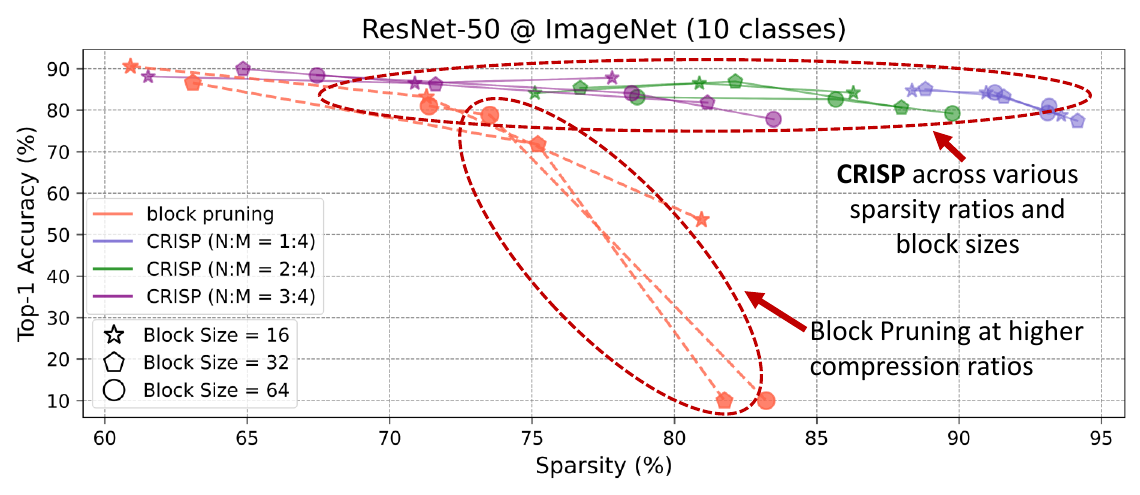}
	\caption{CRISP against block sparsity on ImageNet dataset}
	\label{fig:acc_setup}
  \vspace{-3mm}
\end{figure}

An important advancement in structured pruning research is the introduction of fine-grained 2:4 structured sparsity \cite{Mishra2021AcceleratingSD}, which selectively retains two non-zero values out of every four consecutive tensor values, as shown in Fig. \ref{fig:sparsity} (left). This particular form of sparsity has gained support from NVIDIA GPU \cite{stc}, resulting in theoretical speedup gains of up to 2$\times$ compared to the dense counterparts. The fine-grained 2:4 sparsity pattern limits the sparsity ratio, allowing only 50\% of the total model capacity to be sparse. Nevertheless, different models exhibit varying degrees of compressibility, with models like ResNet-50 being heavily over-parameterized and thus prunable with ease, in contrast to more compact models such as MobileNetV2, as shown in Fig. \ref{fig:diff_nm}. In this work, we aim to explore extending this granularity to varying N:M ratios. 

Nonetheless, this approach still imposes a fixed per-layer pruning ratio on the model, which limits the ability to selectively remove irrelevant layers from the network in a non-uniform manner. As illustrated in Fig. \ref{fig:layerwise_spar}, it becomes evident that specific layers can benefit from more aggressive pruning ($\sim$99\%) compared to others. Additionally, since we are primarily focusing on a small subset of the original class distribution, it becomes feasible to proportionally reduce the model size. However, directly varying the fine-grained N:M ratio per layer introduces increased algorithmic complexity \cite{NEURIPS2021_ad68473a} and requires a multitude of hyperparameters, making it impractical for edge devices.

An alternative approach is coarse-grained block sparsity \cite{Gray2017GPUKF}, where entire blocks within the network are pruned, resulting in a more regular access pattern and improved computational efficiency. Unfortunately, using coarse-grained block sparsity can lead to uneven effects on model accuracy, particularly when critical weights are concentrated within a single block. As demonstrated in Fig. \ref{fig:acc_setup}, block pruning struggles to maintain performance as the sparsity rate reaches over $80$\%. In contrast, our proposed framework \emph{CRISP} maintains decent accuracy ($\sim$85\%) across high compression rates ($>$92\%). 

\emph{CRISP} introduces a novel pruning framework designed for class-aware personalization, leveraging hybrid structured sparsity. We achieve the best of both worlds by inherently integrating fine-grained N:M sparsity with coarse-grained block sparsity to achieve high sparsity levels while preserving model accuracy. 
We utilize a gradient-based class-aware saliency score to carefully select the most significant weight values corresponding to user-preferred classes. Our approach performs global model pruning, ensuring an equal number of non-zero blocks in each row, facilitating efficient hardware implementation.

Our contributions can be summarized as follows:
\begin{itemize}
    \item We introduce \textit{CRISP}, a novel, accurate user-preference-aware pruning framework for model personalization.
    \item We propose a unique hybrid structured sparsity pattern incorporating fine-grained N:M and coarse-grained block sparsity to achieve high model accuracy with superior hardware efficiency.
    \item We introduce an iterative pruning strategy
    based on class-aware saliency estimation, eliminating an equal number of blocks from each row in the weight matrix.
    \item We experiment with three popular models: ResNet-50, VGG-16, and MobileNetV2, achieving comparable accuracy to their dense counterparts across varying user-preferred classes on CIFAR-100 and ImageNet datasets.
    \item We design an accelerator-level hardware design leveraging our hybrid structured sparsity and compare it with state-of-the-art sparse accelerators, achieving up to 14$\times$ speedup and 30$\times$ energy efficiency.
    \end{itemize}

\section{Related Work}
Numerous works have focused on designing user-personalized models tailored to local user preferences on the device. Recent studies, such as CAPTOR and CAPNN \cite{9218741, 9360865}, employ structured pruning techniques in cloud environments, targeting weight values for pruning based on class firing rates to reduce overall memory overhead. Other approaches, like MyML \cite{DBLP:conf/dac/GoyalBD21}, introduce a user-driven channel pruning scheme for edge devices but encounter challenges in maintaining high model accuracy at substantial compression rates. Similarly, OCAP \cite{OCAP} proposes an on-device channel pruning framework leveraging intermediate feature maps and maintains a subset of the training dataset to reduce fine-tuning overhead. Another recent work \cite{chameleon} introduces a memory-efficient approach for continuously updating dynamically evolving models on edge devices. However, CRISP distinguishes itself significantly from existing works by prioritizing hardware efficiency through a novel hybrid structured sparsity pattern. Our pruning framework sustains high accuracy even at extremely high compression rates, presenting a unique solution that seamlessly combines hardware efficiency and accuracy within a framework.

\section{CRISP Pruning Framework}
\textbf{Notation.}
We use the following notations. The neural network parameters are denoted by $f_{\theta}$. For each layer $l$ in the network, with weight tensor $\mathbf{W}_l \in \mathrm{R^{H \times W \times R \times S}}$, where $H$ and $W$ represent the height and width of the convolution kernel, and $R$ and $S$ represent the number of input and output channels, respectively, we reshape the tensor into $\mathbf{W}_l \in \mathrm{R^{HWR \times S}}$. Additionally, we define a binary mask $\mathbf{M}_l \in \mathrm{R^{HWR \times S}}$ to prune each kernel $\mathbf{W}_l$. Here, N and M represent the N:M sparsity ratio, fixed for the entire network.
Our objective is to find the optimal mask $\mathbf{M}_l$ for each layer $l$ while achieving the global compression ratio $\kappa$ with minimal increase in the loss value $\mathcal{L}$ (representing accuracy). 

\subsection{Hybrid Structured Sparsity pattern}
\begin{figure}[!h]
 \vspace{-4mm}
\begin{minipage}[t]{0.6\linewidth}
    \includegraphics[width=\linewidth]{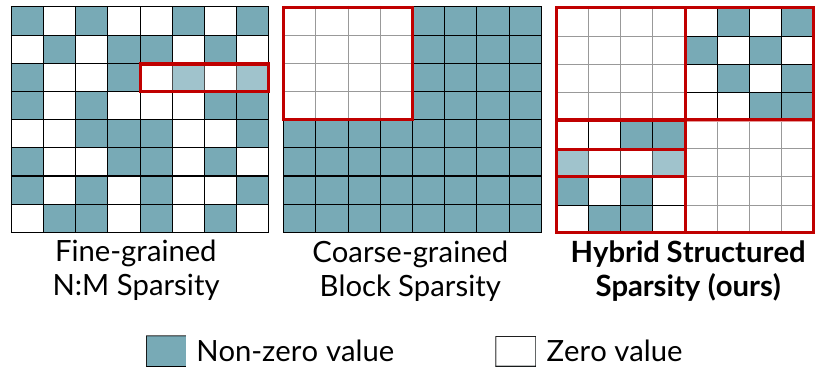}
\end{minipage}%
    \hfill%
\begin{minipage}[t]{0.37\linewidth}
    \includegraphics[width=\linewidth]{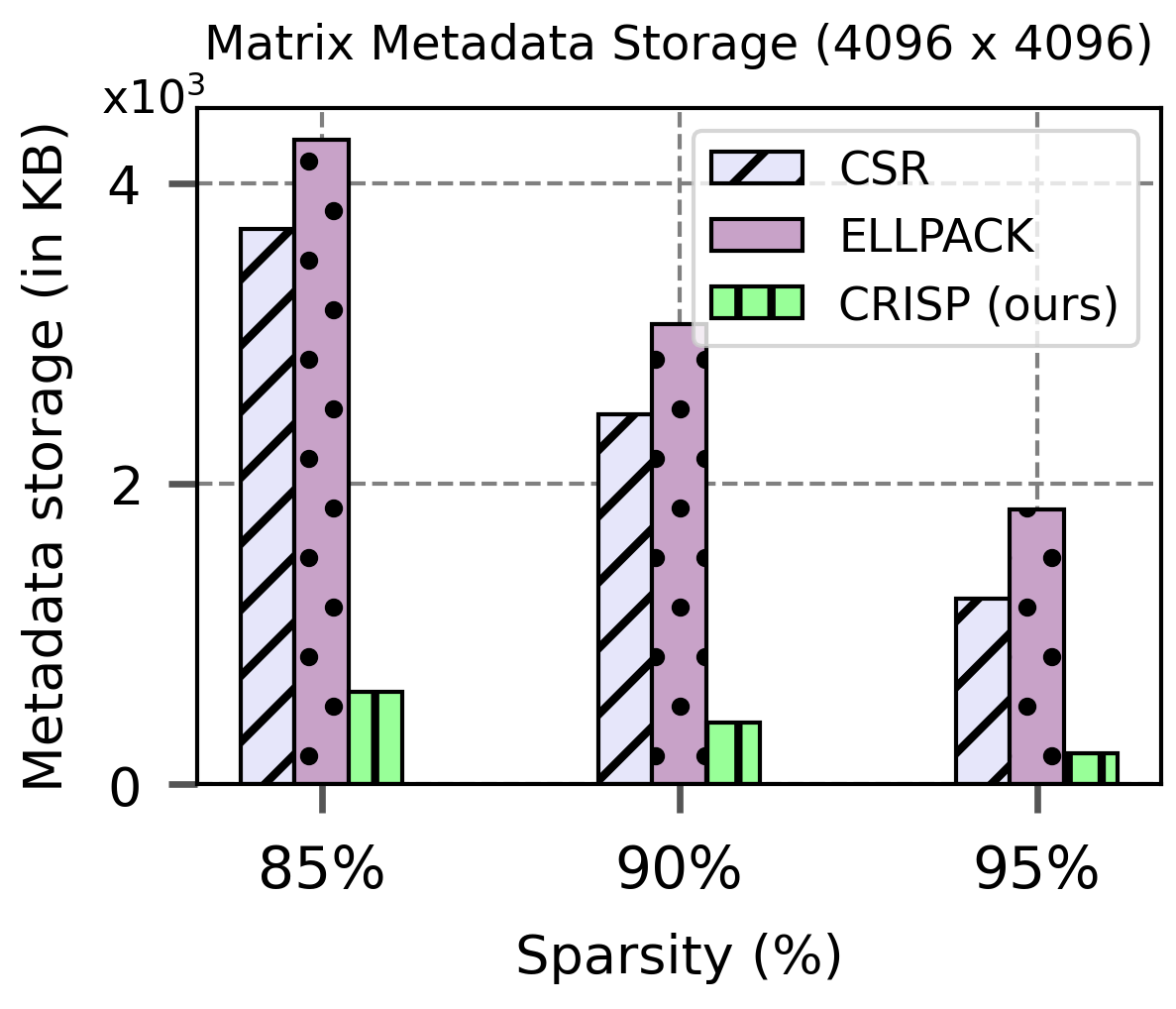}
\end{minipage} 
     \caption{Left: Different structured sparsity patterns, Right: Metadata storage for different formats}
         \label{fig:sparsity}
 \vspace{-3mm}
\end{figure}

The hybrid structured sparsity pattern, as illustrated in Fig. \ref{fig:sparsity} (left), combines the $N$:$M$ fine-grained sparsity pattern, which restricts $N$ non-zero values out of every $M$ consecutive elements and coarse-grained block sparsity, where entire blocks of size $B$ $\times$ $B$ are pruned from the weight matrix. Our proposed sparse pattern takes into account two critical hardware overheads: \textbf{load balancing} and \textbf{metadata storage}. 

\begin{figure*}[!h]
	\centering
\includegraphics[width=0.91\linewidth]{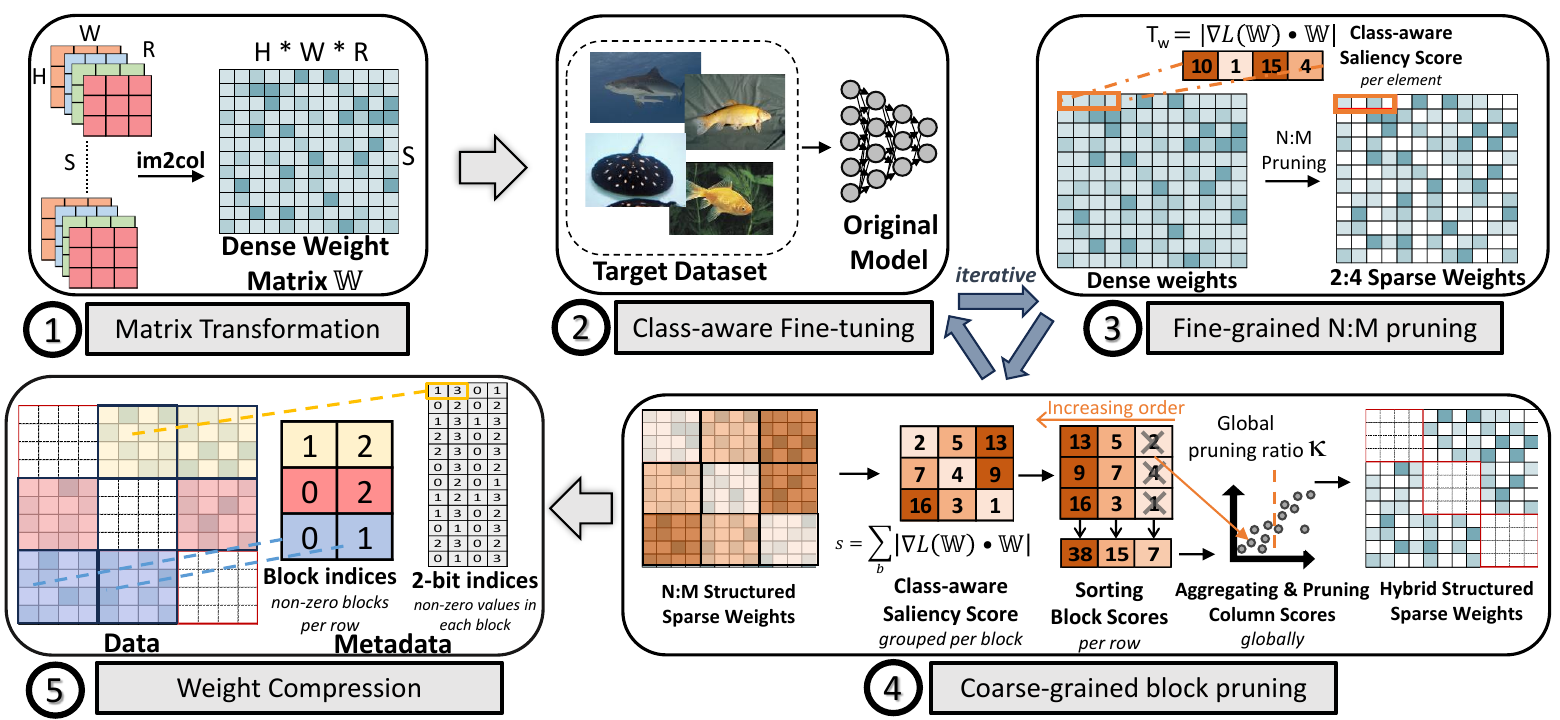}
	\caption{CRISP framework: High-level Overview}
	\label{fig:framework}
   \vspace{-4mm}
\end{figure*}

First, the underlying hardware needs to efficiently identify non-zero values and distribute them evenly to the parallel compute units. For example, NVIDIA-based Sparse Tensor Cores~\cite{stc} support the 2:4 pattern, resulting in a 50\% reduction in weight values with a 2$\times$ speedup. The fixed constraint on the location of non-zero values leads to uniform load balancing, ensuring that every $N$ non-zero values can be easily assigned to $N$ parallel compute units, eliminating ineffectual operations. Several methods leverage formats such as CSR~\cite{csr} and ELLPACK~\cite{ell} to store non-zero values in a compact manner. However, they require zero padding to accommodate a non-uniform number of non-zero elements, resulting in a large memory footprint, complex logic, and increased power consumption. In contrast, our work introduces uniform block pruning such that each row has the same number of pruned blocks across the weight matrix, leading to high workload balancing.

Second, we employ an offset-based coordinate representation to represent non-zero values. We maintain two metadata structures corresponding to N:M sparsity and block sparsity. For block sparsity, we use the Blocked-Ellpack format \cite{blockell} to store column indices of corresponding non-zero blocks in a row-major order. For N:M sparsity, we keep indices of the locations of the non-zero values within a vector of size M. In general, if the original number of columns in the reshaped weight matrix is $K=HWC$, and the number of non-zero columns is $K'$, then the storage requirements for block sparsity can be defined as ($S \times K' \times \lfloor\log_{2}(K'/B)\rfloor) / (B \times B)$ bits, and for N:M sparsity, it can be calculated as $S \times K' \times (N/ M) \times \lfloor\log_{2}(M)\rfloor $ bits. Consequently, the overall average sparsity can be formulated as $1 - ((K'/K) \times (N/M))$. Fig. \ref{fig:sparsity} (right) shows roughly 5$\times$ and 7$\times$ more metadata overhead for CSR and ELLPACK formats, respectively, compared to CRISP.

\subsection{Method Overview}
This section presents a high-level overview of our proposed class-aware pruning framework. Starting with a pre-trained model, we identify the frequently occurring classes within a predefined window as the user-preferred classes denoted as $u_c$. The CRISP algorithm follows a three-step pruning process: (a) class-aware fine-tuning to estimate the class-aware saliency score (CASS), (b) N:M pruning using a straight-through estimator \cite{ste}, and (c) uniform coarse-grained block pruning, as illustrated in Fig. \ref{fig:framework}. These three steps are executed iteratively, gradually increasing the global model sparsity over multiple iterations instead of pruning a large percentage of weights in a single iteration. 

\begin{algorithm}[] 
\caption{CRISP: Class-aware Pruning Framework}
\label{alg:loop}
\begin{algorithmic}[1]
\Require{neural network parameters $f_{\theta}$, Loss function $\ell$, input samples $H$, sparsity level $\kappa$,  iteration steps $n$, $N$, $M$} 
    \For{$p \gets 1$ to $n$}                  \State $f_{\theta}$ $\gets$ {apply fine-grained N:M pruning}      
         \State {$\kappa_p$ $\gets$ $(1 - \frac{N}{M}$) + $\Delta$, update pruning ratio}
     \State Calculate class-aware saliency score $s_j$
 \State  $s_j$ $\gets$  $\sum_{i}$  $|$ ${T_{w}^{i}}$ $|$, for each weight element i in block $j$ 
     \State $S_{j}$ $\gets$ \parbox[t]{200pt}{%
     $\operatorname{Sort([s_j])}$ in increasing order per row in the weight matrix \strut}
     \State   $c_o$ $\gets$ $\sum_j$ ${S_{j}}$, for each block j in column $o$
  \State   $C_o$ $\gets$ \parbox[t]{200pt}{%
  $\operatorname{Sort([c_o])}$ in increasing order for each column $o$ \underline{globally} across the network\strut}
     \State $r$ $\gets$ Select top-$\mathbf{k}$ scores from $C_o$ based on $\kappa_p$
     \State Prune $f_{\theta}^{*}$ $\gets$ $f_{\theta}$ $\odot$ $r$
     \State Fine-tune $f_{\theta}^{*}$ for $\delta$ epochs
    \EndFor    
\end{algorithmic}
\end{algorithm}

\subsection{Pruning Algorithm} 
Algorithm \ref{alg:loop} outlines the pruning algorithm, further illustrated in Fig. \ref{fig:framework}. Given a pre-trained model with transformed weight matrix \circled{1} and user-preferred classes, we initiate the pruning process by fine-tuning the model \circled{2} for a few epochs to estimate the class-aware saliency scores, discussed in Sec.\ref{sec:pruning_metric}. Subsequently, we employ these scores for N:M fine-grained pruning \textit{(Line 2)}  \circled{3}. Here, we extend the Straight-through Estimator (STE) \cite{ste} to derive dense gradients by back-projecting them onto the original weight matrix as approximated gradients. This helps us revisit weights pruned initially due to small or noisy gradient values but may be relevant for the current set of classes. The next step involves coarse-grained block pruning (\textit{Line 4}-\textit{10}) \circled{4}. Here, we partition the weight matrix into a grid of two-dimensional blocks, employing a fixed block size that can vary from 16$\times$16 to 64$\times$64.
For each block $j$, we calculate a per-block pruning score ${s_j}$ (\textit{Line 5}). We independently sort these block scores ${s_j}$ within each row of the two-dimensional block grid \textit{(Line 6)}, as also depicted in Fig. \ref{fig:framework}. To ensure the relative importance of blocks within their respective rows, we aggregate the scores for all blocks within each column $o$ of the two-dimensional block grid, resulting in column-wise scores $c_o$ (\textit{Line 7}).

It is important to note that we maintain an equal number of pruned blocks in each row of the weight matrix.
Arbitrarily varying the number of pruned blocks within each row of the weight matrix poses a significant challenge for workload balancing, as observed in previous works such as \cite{hybridblock}. Consequently, we rank these column-wise scores in ascending order (\textit{Line 8}) across the entire network and select the top-$\mathbf{k}$ columns for pruning, based on the current sparsity target rate (\textit{Line 9}). The blocks are then pruned \textit{(Line 10)} while retaining their original positions within the weight matrix. Lastly, we conduct fine-tuning for $\delta$ epochs \textit{(Line 11)} to recover any accuracy loss incurred during the pruning process.
This iterative process continues until the model reaches the target global compression rate ${\kappa}$.
This iterative approach helps prevent layer collapse \cite{layercollapse}, a phenomenon where specific layers in the network become excessively pruned, leading to the collapse of the layer and, eventually, the entire model. 

\subsection{Pruning Metric} 
\label{sec:pruning_metric}
\textbf{Class-aware Saliency Score}.
For class-aware pruning with personalization, we propose to measure the importance of a parameter by estimating the gradient flow using a small set of input samples. Following \cite{gsm}, we rely on the first-order Taylor series to calculate the change in objective function by eliminating a parameter. For each weight tensor $\mathbf{W}$, we define the pruning metric $T_w$ based on the average gradient accumulated over training samples $H$ based on the user-preferred classes $u_c$: 

\begin{equation}
T_w =   \abs*{\frac{1}{H_{u_c}} \Delta \mathcal{L} (\mathbf{W}) \cdot \mathbf{W}}
\end{equation}

Our pruning criteria $T_w$ serves to identify and eliminate unnecessary parameters based on three distinct relative weight importance criteria.
\textbf{First}, large weight magnitude with small gradients: this criterion targets parameters often deemed less relevant for the given set of user-preferred classes, and their removal does not significantly impact overall model accuracy. \textbf{Second}, small weight magnitude with large gradients: this criterion can frequently be attributed to noise and, hence, is amenable to pruning without substantial adverse effects on the model's accuracy. \textbf{Finally}, both small weight magnitude and small gradients: the third criterion pertains to parameters that are not expected to play a critical role in the model's performance, as their minimal variation for a specific set of user-preferred classes $u_c$ suggests their relative insignificance.

\subsection{Accelerator-level Design}
 We introduce \textit{CRISP-STC}, a hardware accelerator design inspired by NVIDIA's Sparse Tensor Core (STC) \cite{Mishra2021AcceleratingSD}, while customizing it for edge-centric applications with limited resources. Our CRISP-STC design is based on publicly available information and comprises an SMEM-RF-Compute topology featuring 256KB of shared memory (SMEM) and four tensor cores. Each tensor core comprises 64 Multiply-Accumulate (MAC) units and a 1KB Register File. We allocate only a fraction of the SMEM bandwidth to model an edge-centric hardware setup compared to an actual NVIDIA-STC GPU. Each processing element (PE) within CRISP-STC follows SIMD dataflow, akin to GPU cores. While NVIDIA's STC primarily supports the 2:4 pattern, we extend support to accommodate 1:4 and 3:4 fine-grained sparsity patterns within the fabric. 

 \begin{figure}[!h]
	\centering
	\includegraphics[width=0.85\linewidth]{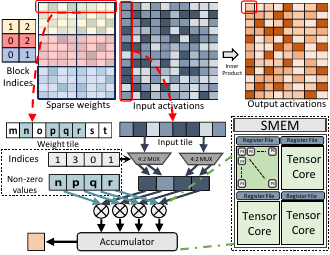}
	\caption{High-level overview of hardware design}
	\label{fig:hardware}
 \vspace{-4mm}
\end{figure}

As illustrated in Fig. \ref{fig:hardware}, CRISP-based sparse matrix multiplication can be divided into three main steps. First, the computation partitions the input and weight matrices into row-major ordered tiles. Block sparsity enables the tracking of non-zero block elements using block indices. Input activations corresponding to non-zero blocks are loaded for computation in the local memory unit (SMEM), taking into account the associated block indices.  

Subsequently, the compressed non-zero blocked input tile undergoes further compression based on the offset-based 2-bit indices metadata. Each non-zero value in the weight matrix is represented by an index within a block of M (in this case, M = 4) values. For example, in Fig. \ref{fig:hardware}, the non-zero value '\textbf{n}' is the second value in its block, denoted by an index of 1. The input tile undergoes processing in the activation selection unit, facilitated by multiplexers (MUXs). Within this unit, an N:M selection process establishes the connection between each block of M weights and their corresponding input values. This operation capitalizes on the pipeline's ability to execute Multiply-Accumulate (MAC) operations for every N out of M weight value. Finally, both the compressed weights and inputs undergo the MAC operation.

\begin{figure*}[!h]
\includegraphics[width=0.97\linewidth]{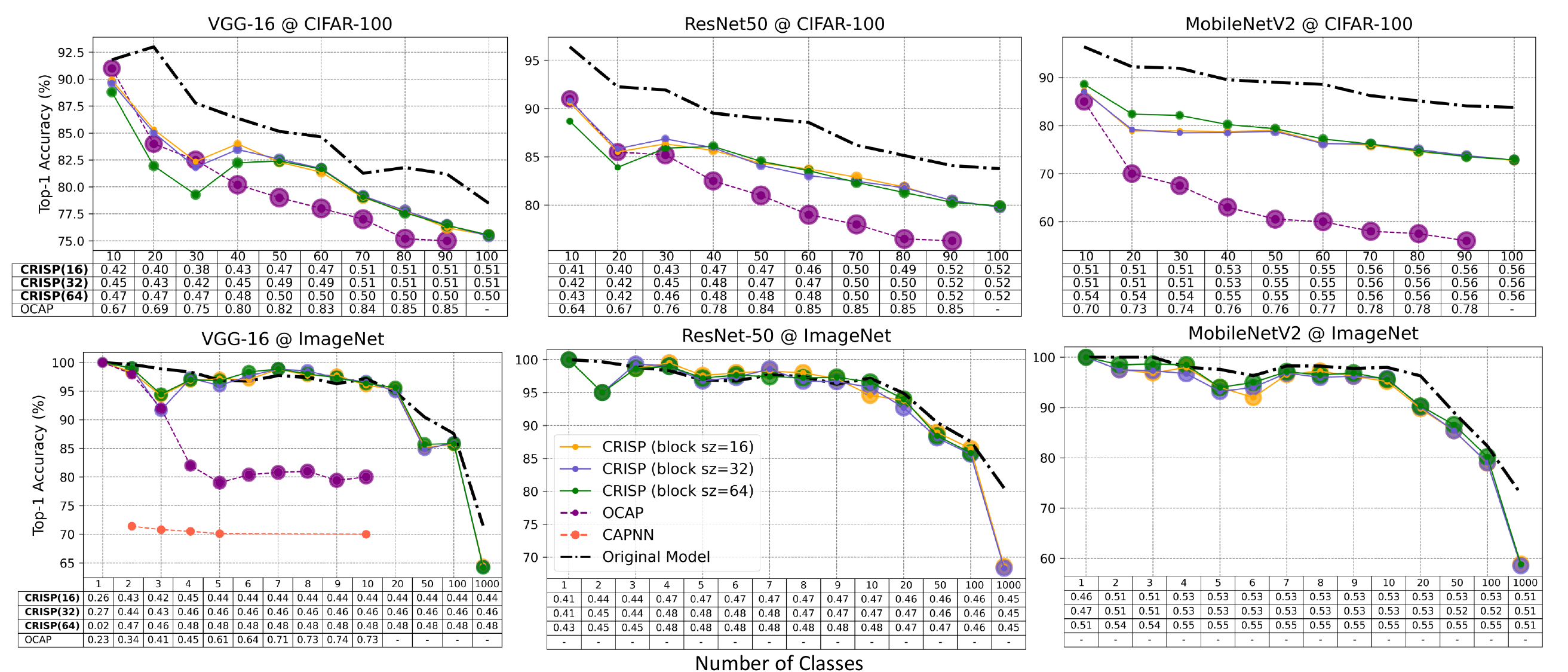}
 \centering
 \caption{Accuracy comparison (averaged over five runs) for different models with varying numbers of classes compared to the baselines. The table at the bottom displays the normalized FLOPs ratio for each method (the smaller, the better).
 } 
\label{fig:accuracy}
  \vspace{-3mm}
\end{figure*}

\begin{figure*}[!h]
\includegraphics[width=0.85\linewidth]{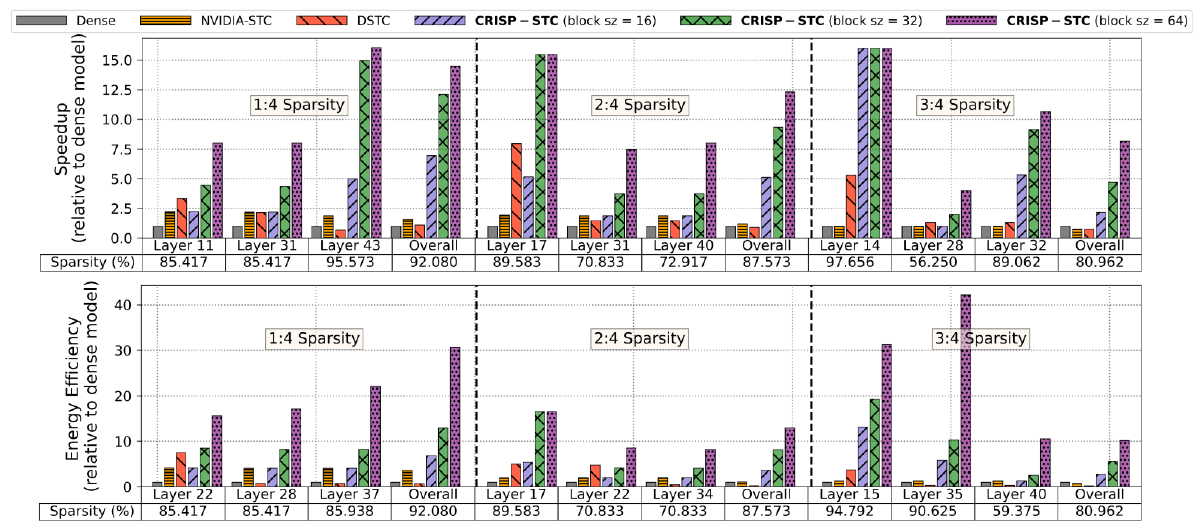}
 \centering
\caption{ResNet-50 layer-wise speedup and energy-efficiency for CRISP-STC compared to NVIDIA-STC\cite{stc} and DSTC\cite{dstc}}
\label{fig:speedup}
  \vspace{-5mm}
\end{figure*}

\section{Experimental Evaluation}

\subsection{Experimental Methodology}

\textbf{Networks and Dataset.}
We evaluate our method using memory-intensive networks on the large-scale ImageNet dataset \cite{ImageNetdataset}: ResNet-50~\cite{res}, VGG-16~\cite{vgg16} and MobileNetv2~\cite{mbv2}. 
For class-aware pruning, we randomly sample 1 to over 1000 classes from the original 1000 classes in the ImageNet dataset
and up to 100 classes from the original 100 classes in the CIFAR-100 dataset. 

\textbf{Training Setup.}
We perform all experiments with the pre-trained ImageNet models. We fine-tune the original dense model on the classes of user interest and report the model accuracy as the upper bound. We also compare against the baselines: OCAP \cite{OCAP} and CAPNN \cite{9218741} with their performance directly obtained from their respective works. We fine-tune all models for $50$ epochs with 
SGD \cite{ruder2016overview} optimizer, the momentum of $0.9$, weight decay of 4e-5, the initial learning rate of $0.1$, a batch size of $32$ and $256$ samples per user-preferred class $u_c$. We also report the normalized \underline{FLOPs ratio} w.r.t the original dense model, which serves as a measure of the compression rate achieved by the pruned model.

\textbf{Hardware Setup.}
We evaluate state-of-the-art sparse accelerators: NVIDIA's Sparse Tensor Core (STC)~\cite{stc}, and Dual-side Sparse Tensor Core (DSTC)~\cite{dstc} with \textit{CRISP-STC} for ResNet-50. We reserve the activation sparsity for DSTC, which can utilize both weight and activation sparsity to $40\%$. As our proposed design and baselines cannot be emulated directly, we use Sparseloop \cite{9923807}, a cycle-accurate simulator for all sparse tensor accelerators. Note that Fig. \ref{fig:hardware} depicts the workflow with a 2:4 sparsity pattern; we adapt the design with an appropriate number of MUXs/multipliers to accommodate other patterns. We report \underline{latency} measurements (cycle counts) and \underline{energy consumption} (in uJ) using the CACTI plugin \cite{6105405}. 

\subsection{Experimental Results}

\textbf{Accuracy Comparison.} We first compare our proposed CRISP pruning method with block pruning on the ImageNet dataset with ten selected user-preferred classes, as shown in Fig. \ref{fig:acc_setup}. Our experiments span varying degrees of granularity, ranging from N:M ratios of 1:4, 2:4, to 3:4, alongside block sizes ranging from 16 to 64. Remarkably, CRISP consistently outperforms block pruning, achieving higher accuracy even at significantly higher sparsity levels, highlighting its robustness and superior performance. 

Next, we comprehensively analyze the accuracy of ResNet-50, VGG-16 and MobileNetv2 for varying levels of user-preferred classes on CIFAR-100 and ImageNet dataset, as shown in Fig. \ref{fig:accuracy} with a fixed 2:4 pattern and varied the global sparsity level ($\kappa$) based on the selected number of user-preferred classes. The four rows at the bottom of Fig. \ref{fig:accuracy} demonstrate the FLOPs ratio for different configurations of CRISP versus OCAP. It is important to note that OCAP's experiments were limited to the CIFAR-100 dataset and VGG-16 on ImageNet, while CAPNN focused exclusively on VGG-16 and ImageNet.  CRISP consistently outperforms the baseline method, achieving higher accuracy with significantly lower FLOP ratios for both the ImageNet and CIFAR-100 datasets. Notably, for VGG-16 on the ImageNet dataset (with ten classes), CRISP maintains over 95\% accuracy with only a 0.44 FLOPs ratio, whereas OCAP achieves $\sim$80\% accuracy with a much higher FLOPs ratio of 0.73. Across all block sizes, we observe a consistent pattern where accuracy experiences a slight drop with an increase in the number of classes. 

\textbf{Hardware Performance.} We analyze layer-wise computation time and energy efficiency of our proposed CRISP pruning method for different fine-grained N:M sparsity (N={1, 2, 3} and M=4) with NVIDIA-STC, DSTC, and the original dense ResNet-50 implementation, as shown in Fig. \ref{fig:speedup}. We evaluate CRISP-STC using various block sizes, including 16$\times$16, 32$\times$32, and 64$\times$64,
with a global sparsity in the range of 80\%--90\%. We also provide performance comparisons for a  representative set of selected layers, along with their sparsity rates. 

CRISP-STC consistently outperforms all three baselines, achieving approximately 7--14$\times$, 5--12$\times$, and 2--8$\times$ speedup for 1:4, 2:4, and 3:4 fine-grained sparsity, respectively. In contrast, NVIDIA-STC can only achieve up to 2$\times$ speedup, even for the 1:4 sparsity pattern, primarily due to its poor utilization rate. Interestingly, DSTC can achieve roughly 3--8$\times$ speedup for initial layers with larger input dimensions, yet fails to maintain similar performance for later layers, where data movement becomes the bottleneck, resulting in poor overall latency, as also discussed in their work \cite{dstc}.  
Among the different block sizes, a block size of 64 demonstrates superior performance due to its better compression ratio and more efficient load balancing, resulting in more streamlined computations. 
In terms of energy consumption, CRISP achieves up to 30$\times$ energy efficiency compared to the dense model. Once again, a block size of 64 exhibits dominant energy efficiency over the other block sizes. DSTC spends a significant amount of energy due to its complex dataflow, while NVIDIA-STC can only achieve up to 2$\times$ energy efficiency.
\newline
\section{Conclusion}
This work proposes \emph{CRISP}, a resource-efficient pruning framework for class-aware personalization. CRISP introduces a novel structured sparsity pattern consisting of coarse-grained block and fine-grained N:M sparsity. The framework leverages an iterative gradient-preserving pruning strategy to preserve the model weights corresponding to user-specific classes. Our evaluations show CRISP achieves accuracy comparable to their dense counterparts for ResNet-50, VGG-16, and MobileNetV2, with over 90\% sparsity across all three models. Moreover, the resulting pruned models exhibit up to 14$\times$ latency and 30$\times$ energy efficiency compared to existing baselines. We plan to extend these results to transformer-based architectures.

\section{Acknowledgment}
This work is partially supported by the
National Research Foundation, Singapore, under its Competitive Research Programme Award NRF-CRP23-2019-0003 and Singapore Ministry of Education Academic
Research Fund T1 251RES1905. 

\bibliographystyle{unsrt}
\bibliography{date}

\end{document}